\pdfoutput=1

\documentclass[11pt]{article}

\usepackage[]{emnlp2022}

\usepackage{times}
\usepackage{latexsym}

\usepackage[T1]{fontenc}

\usepackage[utf8]{inputenc}

\usepackage{microtype}

\usepackage{amsmath}
\usepackage[linesnumbered,ruled]{algorithm2e}
\usepackage{subfig}

\usepackage[labelfont=bf]{caption}
\usepackage{graphicx}
\usepackage{comment}
\usepackage{mathtools}

\usepackage{multirow}
\usepackage{multicol}
\usepackage{booktabs}

\title{SLATE: A Sequence Labeling Approach for Task Extraction from Free-form Inked Content}

\author{Apurva Gandhi$^{1*}$, Ryan Serrao$^{2}$, Biyi Fang$^{1}$, \\ {\bf Gilbert Antonius$^{1}$, Jenna Hong$^{1}$, Tra My Nguyen$^{3}$, Sheng Yi$^{4}$, } \\ {\bf Ehi Nosakhare$^{1}$, Irene Shaffer$^{1}$, Soundararajan Srinivasan$^{1}$, Vivek Gupta$^{5}$}\\ Microsoft \\ \{$^{1}$firstname.lastname, $^2$ryserrao, $^{4}$shengyi, $^{3}$nguyenm, $^{5}$vivgupt\}@microsoft.com}

\begin{document}
\maketitle

\begin{abstract}
We present SLATE, a sequence labeling approach for extracting tasks from free-form content such as digitally handwritten (or "inked") notes on a virtual whiteboard. Our approach allows us to create a single, low-latency model to simultaneously perform sentence segmentation and classification of these sentences into task/non-task sentences. SLATE greatly outperforms a baseline two-model (sentence segmentation followed by classification model) approach, achieving a task F1 score of 84.4\%, a sentence segmentation (boundary similarity) score of 88.4\% and three times lower latency compared to the baseline. Furthermore, we provide insights into tackling challenges of performing NLP on the inking domain. We release our code and dataset for this novel task. 
\end{abstract}

\section{Introduction}\label{introduction}
\renewcommand*{\thefootnote}{\fnsymbol{footnote}}
\footnotetext{$^*$Corresponding author.}
\renewcommand*{\thefootnote}{\arabic{footnote}}

The shift to remote and hybrid working styles due to COVID-19 has led to a large increase in virtual meetings.  It has become increasingly important for participants to express themselves and brainstorm as naturally and effortlessly as possible, leading to an opportunity to extract entities from the large amounts of content created in these meetings. A natural entity of interest is a task created by a participant during the meeting which can be assigned to an individual to complete afterwards.


While past works have investigated extracting tasks from typed content such as emails \cite{bennett2005detecting, wang2019context}, there has been less focus on task extraction from more \textit{free-form} content such as digitally handwritten (or \textit{inked}) content on a virtual whiteboard or spoken content in a meeting. Extracting tasks from free-form content is challenging as it is often not as well-structured (e.g., poor grammar, inconsistent/lack of punctuation, typos, etc.) Furthermore, since this content first needs to be converted to text (e.g., through a handwriting recognition model for ink or ASR for speech), downstream NLP models must be robust to errors made by the recognition models.

\begin{figure}[t!]
    \centering
    \includegraphics[width=\linewidth]{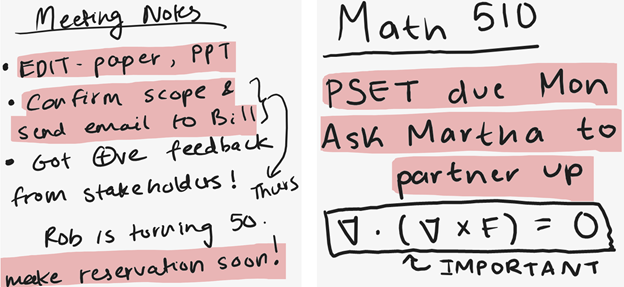}
    \caption{Examples of inked content. Task sentences are highlighted in red. Note the free-form style of the content, containing lists, a paragraph and annotations.}
    \label{fig_sample_board}
    \vspace{-5mm}
\end{figure}

Moreover, as we discuss in Section~\ref{related_work}, past approaches for task extraction from typed content assume text to already be segmented into sentences and focus on building a separate sentence-level task classification model. We cannot make this assumption for our scenario since automatic sentence segmentation models trained on typed content are unlikely to generalize well to the inconsistent capitalization and punctuation in free-form content \cite{stevenson2000experiments, rehbein2020improving}. Furthermore, separating classification and sentence segmentation creates a latency-challenge: Classifying each sentence separately can cause the latency on CPU to scale linearly with the number of sentences, making real-time extraction challenging.

In our work, we address these challenges with our proposed approach SLATE – A Sequence Labeling Approach to Task Extraction from free-form content. Particularly, we apply this approach to inked content, such as that in Fig.~\ref{fig_sample_board}, and call this application as Ink-SLATE. Our contributions are: 
\begin{itemize}
    \vspace{-2mm}
    \item We create a single, low-latency sequence labeling model to simultaneously perform sentence segmentation and task sentence classification for inked content in a document. 
    \vspace{-2mm}
    \item We leverage ink document layout features to overcome inking domain challenges such as the lack of punctuation and capitalization. 
    \vspace{-2mm}
    \item We discuss a custom evaluation procedure suitable for the problem of extracting task sentences from free-form content and benchmark our SLATE approach against a baseline two-model approach. 
    \vspace{-2mm}
    \item We compile a novel dataset for task extraction from ink document text. We release both our code and dataset in our linked repository.\footnote{Dataset, code, and additional details available at: \href{https://github.com/SLATEAuthors/SLATE}{https://github.com/SLATEAuthors/SLATE}}
    \vspace{-2mm}
\end{itemize}

\section{Related Work}\label{related_work}
Past work on analyzing digitally inked content has dominantly fallen into either the category of handwriting recognition \cite{multilanghandwriting, gericke2012handwriting} or document layout analysis -- grouping words into document lines \cite{linegrouping}, grouping document lines into blocks of spatially related text \cite{block_grouping}, or detecting indentation \cite{list_detection}. Our work differs from these past works as it analyzes the semantics of the inked content rather than the layout. Furthermore, in an ink-analysis pipeline our work would sit downstream to handwriting recognition and layout analysis rather than replace them; we use, as input to our task extraction system, both recognized text and layout information extracted from the inked document. To our knowledge, our work is the first to tackle both sentence segmentation and task extraction for inked content.

Past work on task extraction has mainly been applied to typed content such as emails \cite{bennett2005detecting, wang2019context}. Other than our different domain of \textit{inked content}, our work also differs in approach from these past works.  Particularly, these works assume input text to be already segmented into sentences and focus only on building a classification model to classify these sentences into tasks/non-tasks. Thus, implicitly the task extraction systems in these works rely on a two-model approach: A sentence segmentation model to produce sentence candidates, followed by a classification model to classify sentences as tasks/non-tasks or different sub-categories of tasks. Our approach, on the other hand, uses a single model to simultaneously perform both sentence segmentation and classification for inked content.

Sequence labeling is an NLP approach that predicts a label for each token within a sequence, rather than a label for the whole sequence. Applications of sequence labeling have traditionally included named entity recognition (NER), part-of-speech (POS) tagging, text segmentation, etc. In our work, we reframe the typically two-stage problem of task sentence extraction from documents as a single-stage sequence labeling problem. Since our approach uses a single, shared model to both segment and classify text, it can be thought of as a form of multi-task learning. Multi-task learning has shown to be data-efficient and less prone to overfitting to any single task \cite{multitask_learning}.

Most previous works on sequence labeling use Bi-LSTM and CRF layers in their models \cite{seq_label_survey, chen2020seqvat}. We instead use a RoBERTa architecture, following the recent success of fine-tuning pretrained transformer LMs for data-constrained NLP. \cite{wolf2020transformers}.  

\vspace{-2mm}
\section{Our Approach}
\vspace{-2mm}
\subsection{Dataset}\label{dataset}
Our dataset consists of 200 vendor-created ink documents. To generate these documents, the vendors were provided various example templates with different content (to-do lists, recipes, brainstorms, general notes, etc.) that contain tasks and non-tasks written in various styles (single sentences, paragraphs, lists, diagrams, etc.) For additional diversity, the vendor employed 50+ different donors from different genders and age groups, and with various writing habits (e.g., left/right-handed).

After obtaining these ink documents, we passed them through a handwriting recognition engine (with 9.8\% word error rate) and document layout analysis engine similar to ones referred to in Section~\ref{related_work} or to publicly available APIs~\cite{April_myscript_api}. The result of this is 200 document text blocks and associated layout metadata (line breaks, bullets, etc.; see Section~\ref{domain_adaptation}). The layout analysis also groups spatially related regions of text into separate blocks (similar to \cite{block_grouping}), which we refer to as \textit{writing regions}. We then split these writing regions between 6 annotators who performed two kinds of annotations: (1) Inserting sentence boundaries for sentence segmentation; (2) Annotating each sentence segment as a task/non-task.

Furthermore, the annotators also specially marked certain sentences which were only tasks or non-tasks in the context of the neighboring sentences. An example is a sentence with many misrecognized words, making it incomprehensible in isolation; nevertheless, in the context of a to-do list it may become apparent that the sentence is a task. Additional examples of tasks/non-tasks due to context can be found in Appendix~\ref{ink_document_examples}. Table~\ref{data_stats_table} shows dataset split and annotation statistics. For annotation consistency, the annotators were provided a comprehensive annotation guide with example categories of sentences to be labeled as tasks and non-tasks. We release this guide in our linked repository. Since our task is novel, for reproducibility and to support future research, we also make the document texts, layout metadata, and task/sentence annotations available in our repository\footnotemark[1].

Next, we share domain-specific challenges:
\vspace{-2mm}
\paragraph{Digital handwriting is often misrecognized:} We rely on existing handwriting recognition models to convert handwritten strokes to text. Since handwriting is often messy and diverse, recognized text that is inputted to our task extraction model is often ridden with typos and non-sensical words. 
\vspace{-2mm}
\paragraph{Inked content is often overly concise:} Users are generally not verbose when inking. Rather, they distil their content to important keywords, phrases, acronyms, phrases, etc. This concise style often lacks proper punctuation and grammar, making NLP tasks such as sentence segmentation to find the boundaries of task/non-task sentences quite challenging. For example, the first inked bullet in Fig.~\ref{fig_sample_board} lacks grammar, punctuation and is a list of keywords rather than a proper sentence. Furthermore, the lack of verbosity and the use of esoteric short-hands and acronyms make it even more challenging for a model to understand the meaning of the text. The third bullet in Fig.~\ref{fig_sample_board} shows such a shorthand -- using `$\bigoplus$ve' instead of `positive.'
\vspace{-2mm}
\paragraph{Ink users want to write in an unrestricted, free-form manner:} Inking is conducive to brainstorming. Since people brainstorm tasks in various formats (to-do list, paragraphs, diagrams, mix of these styles, etc.), NLP systems built to analyze inked content must be able to handle this diversity. Fig.~\ref{fig_sample_board} shows an example of the free-form nature of inked content (more examples in Appendix~\ref{ink_document_examples}).

\begin{table}[h!]
\centering
\small
\scalebox{0.78}{
\begin{tabular}{@{}lll@{}}
\toprule
\multicolumn{1}{c}{\multirow{3}{*}{Content}} & \multicolumn{2}{c}{Count} \\
\cmidrule(lr){2-3} 
& \multirow{2}{2.2cm}{Train set} & \multirow{2}{2.2cm}{Test set} \\
\\
\midrule
Ink documents & 124 & 83 \\
\midrule
Sentences & 2496 & 1416 \\
\midrule
Task sentences & 704 & 440 \\
\midrule
Non-task sentences & 1522 & 857 \\
\midrule
\begin{tabular}[c]{@{}l@{}}Task sentences\\ due to context\end{tabular} & 173 & 54 \\
\midrule
\begin{tabular}[c]{@{}l@{}}Non-task sentences\\ due to context\end{tabular} & 97 & 65 \\
\bottomrule
\end{tabular}
}
\caption{Dataset statistics after annotation process.}
\label{data_stats_table}
\vspace{-5mm}
\end{table}

\subsection{Sequence Labeling Approach}\label{sequence_labeling_approach}
\begin{figure*}
    \centering
    \includegraphics[width=\linewidth]{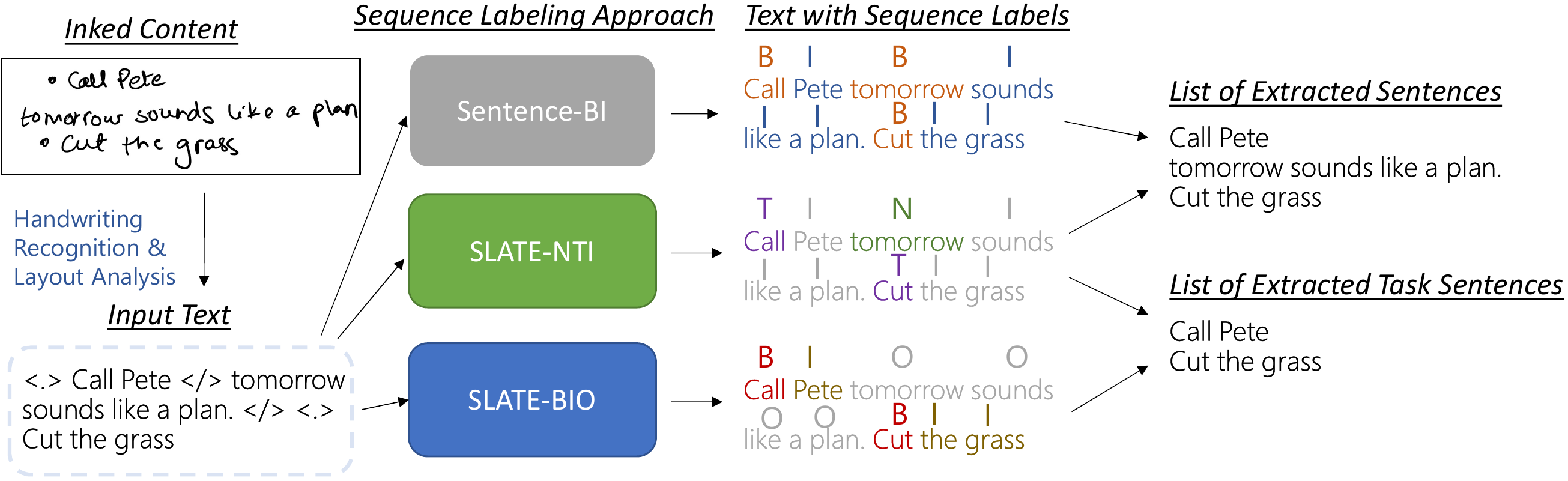}
    \caption{Illustration of the various sequence labeling configurations and how they are used to extract sentences and tasks from inked content.}
    \label{fig_seq_label}
    \vspace{-5mm}
\end{figure*}
At the heart of SLATE is sequence labeling. A sequence labeling approach treats the input document text as a sequence of tokens (or sub-words). It classifies each token as being part of one of a predefined set of classes. To extract our desired entities (e.g., sentences for sentence segmentation or task sentences for task extraction) we post-process the sequence of tokens according to their predicted class labels. A particular \textit{sequence labeling scheme} determines the set of classes and the logic to post-process the predicted token-level class labels for entity extraction. In this work, we define and try three different sequence labeling schemes, described in the sections below. As will be discussed in Section~\ref{SLATE-NTI}, sequence labeling is key to letting us simultaneously perform sentence segmentation and task sentence classification with a single model.  
\vspace{-2mm}
\subsubsection{Sentence-BI Labeling Scheme for Sentence Segmentation}\label{sentence_BI}
The sentence-BI labeling scheme is used for sentence segmentation and is similar to schemes adopted by past works in sentence segmentation \cite{rehbein2020improving, le2020sequence}. In this labeling scheme, tokens are assigned one of two labels: (B) - Beginning of Sentence; (I) - Inside of Sentence.

After the sequence labeling model classifies each token, we aggregate token-level class labels to word-level labels. This is done to make sure that we do not split sentences in the middle of words. The rule used for this aggregation is described in Algorithm \ref{algorithm1} of Appendix \ref{aggregation_rules}. Once we have predicted word-level labels, the words labeled as `B' indicate the beginning of a new sentence, giving us the predicted sentence boundaries for sentence segmentation as shown in the top row of Fig.~\ref{fig_seq_label}.

Since a model trained with the sentence-BI labeling scheme is only useful for sentence segmentation, for our task extraction scenario, we need an additional classification model to classify the segmented sentences into tasks/non-tasks. This two-model approach is precisely what we use as our baseline, described in Section \ref{baseline_section}.

\subsubsection{SLATE-BIO Labeling Scheme for Task Extraction}\label{SLATE-BIO}
The SLATE-BIO labeling scheme is used to extract \textit{task} sentences from the input text. It assigns one of the following three labels to each token: (B) - Beginning of Task Sentence; (I) - Inside of Task Sentence; (O) - Outside of Task Sentence. BIO labeling schemes have commonly been used for NER and text chunking tasks \cite{conllchunking, ramshaw1999textchunking}. In our work, we adapt it for task sentence extraction.

Similar to sentence-BI, we aggregate predicted token-level labels to word-level labels (described in Algorithm \ref{algorithm2} of Appendix \ref{aggregation_rules}). Once we have the predicted word labels, a sequence of labels that starts with a `B' and ends in zero or more `I' labels indicates a task sentence. Task extraction with SLATE-BIO is illustrated at the bottom of Fig.~\ref{fig_seq_label}.

\subsubsection{SLATE-NTI Labeling Scheme for Task and Sentence Extraction}\label{SLATE-NTI}
A disadvantage of SLATE-BIO is that while it finds the boundaries surrounding task sentences, it cannot be used for sentence segmentation as it does not find the boundaries between a contiguous block of non-task sentences. For this, we propose the SLATE-NTI labeling scheme to simultaneously train a sequence labeling model for both task sentence extraction and sentence segmentation (a form of multi-task learning). This scheme assigns one of the following three labels to each token: (N) - Beginning of a Non-Task sentence; (T) - Beginning of a Task Sentence; (I) - Inside of a Sentence. 

Similar to the other schemes, we aggregate predicted token-level labels to word-level labels (Algorithm \ref{algorithm3} of Appendix \ref{aggregation_rules}). Once we have the predicted word labels, a sequence of word labels that starts with a `T' and ends in zero or more `I' labels indicates a task sentence, whereas a sequence that starts with an `N' and ends in zero or more `I' labels indicates a non-task sentence. Sentence segmentation and task sentence extraction using SLATE-NTI is illustrated in the middle row of Fig.~\ref{fig_seq_label}.

\subsubsection{Tackling Inking Peculiarities using Document Layout Metadata}\label{domain_adaptation}
As mentioned in Section \ref{dataset}, inked content is often written in a casual style, lacking punctuation, proper grammar, verbosity, etc. Furthermore, upstream components like handwriting recognition can introduce errors. This makes modeling especially difficult as standard sentence segmentation relies on punctuation and capitalization to determine sentence boundaries. Similarly, misspelled words, acronyms and improper grammar make it difficult for a model to make sense of the sentence's meaning and thus classify it. To compensate for these peculiarities of inked content, we supplement the input to our sequence labeling model with document layout metadata. Particularly, we add the following to our model input.

\textit{Line breaks} indicate where a document line ends and a new one begins. While line breaks do not correspond exactly to sentence boundaries, we expect there to be strong correlation between their positions, providing a useful signal to the model for sentence segmentation purposes. We use the `</>' token to indicate a line break in text.

\textit{Bullets} are used to indicate the start of list items. People tend to write tasks in the form of to-do lists and thus it is common for tasks to be bulleted. Furthermore, bullets almost always indicate the start of a new sentence. Thus, we expect bullets to provide useful signal for both sentence segmentation and task classification. We use the `<.>' token to indicate a bullet in text.

The left side of Fig.~\ref{fig_seq_label} shows how we add line breaks and bullets to the model input. 

\subsection{Baseline Approach}\label{baseline_section}
Our baseline approach is a two-model approach where we first train a sentence segmentation model that takes as input the document text and outputs sentence boundaries. In our work, to build the sentence segmentation model, we use sequence labeling with the Sentence-BI labeling scheme as discussed in Section \ref{sentence_BI}. We then train a separate sentence classification model that takes as input a sentence and outputs a task/non-task label. 

\subsection{Model Architecture}
For each of the models we train, we fine-tune a pretrained RoBERTa \cite{liu2019roberta} encoder implemented in the HuggingFace transformers library \cite{wolf2020transformers}. For the sequence labeling (SLATE-NTI, SLATE-BIO, and Sentence-BI) approaches, we add classification heads for each input token, to obtain the token-level sequence labels. For the classification model in the baseline approach, we use only a single classification head instead, for classifying one sentence at a time. Leveraging a pretrained transformer model allows us to obtain good performance even with our relatively small training set. For training and implementation details, please refer to Appendix \ref{training_details}.

\subsection{Evaluation Procedure}
\begin{figure}
    \centering
    \includegraphics[width=0.9\linewidth]{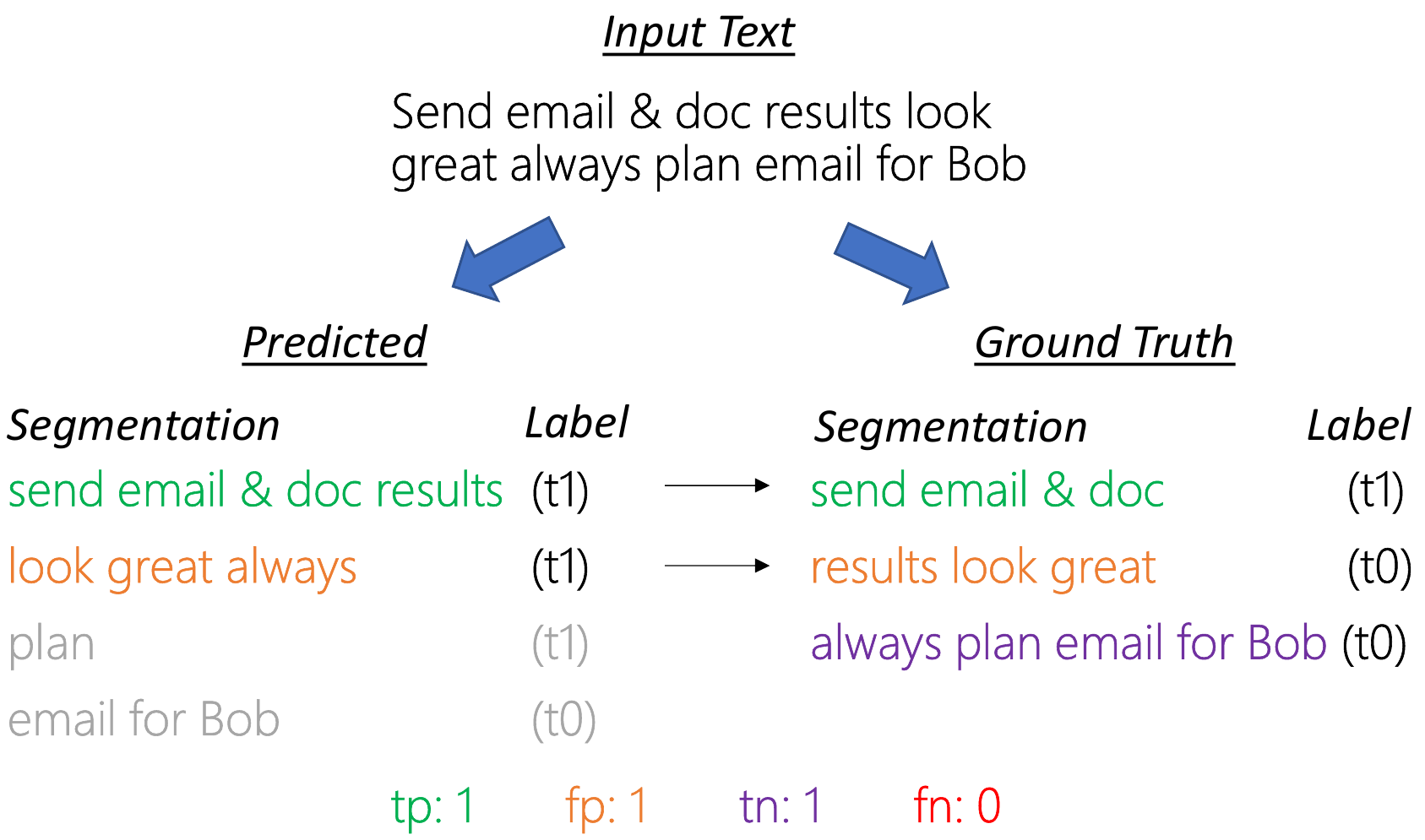}
    \caption{Example of calculating the number of true positives (tp), false positives (fp), true negatives (tn) and false negatives (fn) for the given input text, predictions and ground truth annotations. The t0/t1 labels are used as abbreviations for non-task/task labels.}
    \label{fig_intermediate_metrics}
    \vspace{-4mm}
\end{figure}

A challenge of evaluating SLATE is that since it performs sentence segmentation and classification \textit{jointly}, it is ambiguous how to evaluate its task classification performance. Particularly, since the predicted sentence segments may not match the ground truth sentence segments, it is unclear how to compare their task/non-task labels. For example, consider Fig.~\ref{fig_intermediate_metrics}. At the top, this figure shows sample input text for our task extraction system. The lower left shows the predicted annotation (sentence segmentation and task classification labels) while the lower right shows the ground truth annotation. The predicted task ``send email \& doc results" has words from two ground truth sentences -- ``send email \& doc" and ``results look great." It is unclear which ground truth sentence we should compare its classification label with. Thus, without an explicit matching from predicted task segments to ground truth segments, it is ambiguous how to compare the labels of predicted and ground truth sentences. In our work, we use a bipartite graph matching algorithm to construct such an explicit matching using IOU similarity as edge weights between predicted and ground-truth sentences (Section~\ref{matching_section}). The result of this procedure is an explicit matching, allowing us to port typical classification metrics to our scenario. For example, Fig.~\ref{fig_intermediate_metrics} shows how we can calculate true/false positives/negatives, from which further classification metrics (e.g., accuracy, F1, etc.) can be computed. Additional discussion on our evaluation procedure and comparison to standard NER metrics is provided in Appendix~\ref{evaluation appendix}.

For evaluating the sentence segmentation performance of SLATE, we use the \textit{boundary similarity (B)} sentence segmentation metric introduced in \cite{fournier2013evaluating}. Concisely, \textit{B} penalizes a predicted segmentation based on the number of edits required to transform the predicted segmentation to the ground truth segmentation. Near boundary misses are penalized less compared to full misses/additions. \textit{B} is a score from 0-1 where a higher score represents a better predicted segmentation and a 1 represents a perfect segmentation. More metric details are in Appendix~\ref{app_boundary_sim}.
    
In our application, since the segmentation quality of extracted task sentences matters more than that of non-task sentences, we also compute a modified version of \textit{B} which we call the \textit{true positive boundary similarity ($B_{tp}$)}. The formula to compute $B_{tp}$ is the same as Equation \ref{eq_boundary_similarity} (Appendix~\ref{app_boundary_sim}) except that in the segmentations that we compare, we only include the boundaries of true positive tasks.

\subsubsection{Matching Predicted Task Sentences to Ground Truth Sentences}\label{matching_section}

\begin{figure}[t!]
\centering
\subfloat[Step 1 of the matching process constructs a complete weighted bipartite graph between the sets of predicted task sentences and ground truth sentences. The edge weights represent the IOU similarity between sentences, calculated on their respective sets of word indices (purple). The light gray edges have zero edge weights.]{
\includegraphics[width=1.0\linewidth]{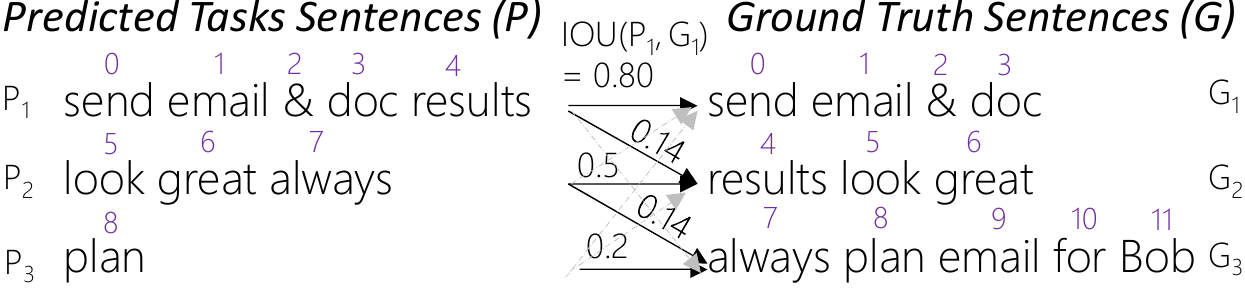}
}

\subfloat[Step 2 of the matching process finds the maximum weight full matching for the constructed graph.]{
\includegraphics[width=1.0\linewidth]{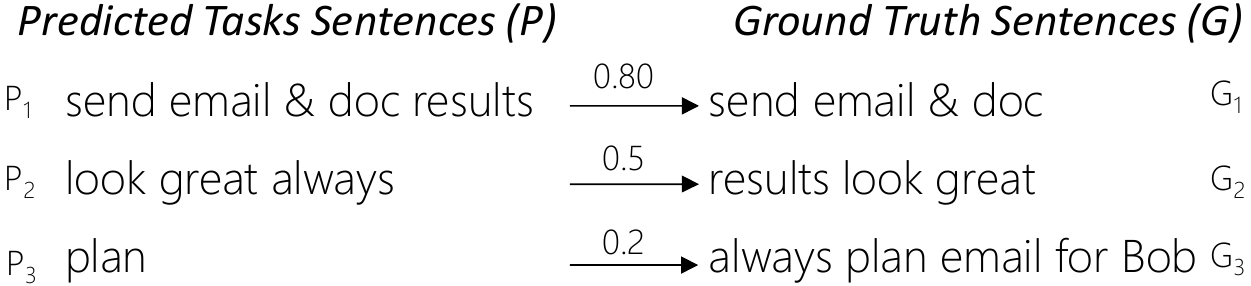}
}

\subfloat[Step 3 of the matching process prunes out edges that do not meet a minimum similarity threshold.]{
\includegraphics[width=1.0\linewidth]{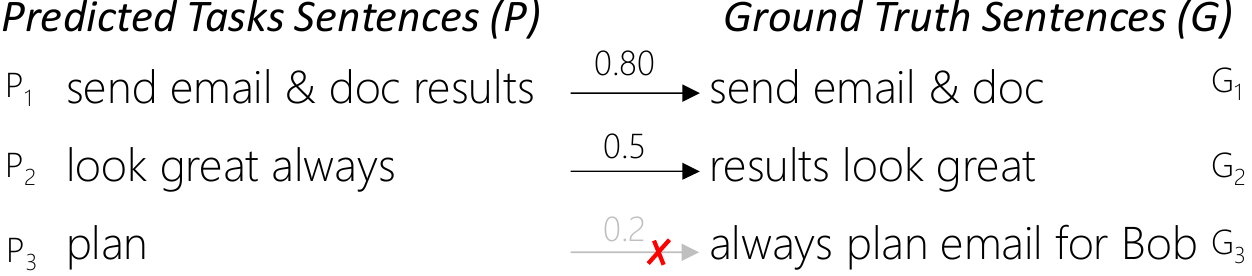}
}
\caption{Illustration of the procedure used to match predicted task sentences to ground truth sentences.}
\vspace{-4mm}
\label{fig_matching_process}
\end{figure}

In this section, we describe the procedure used to obtain a matching between predicted task sentences and ground truth sentences. Let $D$ represent the document text provided to the model to perform inference on. $D$ is then a sequence of words $w_i$ where $index(w_i)$ represents the positional index of $w_i$ in $D$. Let $G$ be a partition of $D$ corresponding to the ground truth sentences (task and non-task) in $D$, i.e, each element $G_i \in G$ is a set of words $w_j \in D$ representing a sentence in the ground truth segmentation of $D$. After performing inference with our model on $D$ we observe the set $P$ corresponding to the set of predicted task sentences, i.e, each element $P_i \in P$ is a set of words $w_j \in D$ representing a predicted task sentence. With this notation, we are now ready to describe the steps of the matching procedure: 
\begin{enumerate}
    \vspace{-2mm}
    \item \textbf{Construct a complete weighted bipartite graph between the sets $P$ and $G$ where each element (sentence) in the sets is a node and edge weights represent similarity between the node sentences.} A complete bipartite graph between sets $P$ and $G$ is one where every pair of nodes from differing sets have an edge but no pair of nodes from the same set has an edge. We use the Intersection over Union (IOU) between sentences in the graph to measure similarity. In our scenario, we define IOU as follows:\footnote{Using $P_{i-ind} \, \cap \,  G_{j-ind}$ in the numerator of the definition of $IOU(P_i, G_j)$ instead of simply $P_{i} \, \cap \, G_{j}$ is important to avoid spurious matches when the same words may be repeated more than once in $D$.}
    \vspace{-3mm}
    \begin{multline}
    IOU(P_i, G_j) = \frac{|P_{i-ind} \cap G_{j-ind}|}{|P_i \cup G_j|}\\
    \text{where} \ S_{-ind} \coloneqq \{index(w_l)~|~w_l \in S\}.
    \label{eq_IOU}
    \end{multline}
    \vspace{-8mm}
    \item \textbf{Find the maximum weight full matching $M$ for the bipartite graph constructed above.} We desire a matching between the sets $P$ and $G$ to maximize the overall similarity between matched sentences. Let $C = ((P, G), E)$ represent the weighted complete bipartite graph we constructed in first step, where the set $E$ represents the set of edges in the graph and $weight(e)$ for $e \in E$ represents the IOU similarity score between the nodes that $e$ connects. We construct a matching $M \subseteq E$ such that each node in the graph is included in at most one edge in $M$ and so that $|M| = min(|P|, |G|)$. When $|P| = |G|$, this is commonly known as a perfect matching. Since in our scenario we allow $|P|$ and $|G|$ to differ, we follow \citet{karp1980algorithm} and refer to this as a \textit{full} matching. Furthermore, we choose the edges in $M$ so as to maximize $\sum_{e \in M} weight(e)$, making $M$ the maximum weight full matching for graph $C$. Essentially, $M$ chooses the edges between predicted task sentences and ground truth sentences that maximizes the overall similarity between matched sentences. Also, note that the full matching allows at most one predicted segment to be matched to a ground truth sentence, preventing overcounting in the case where there are more than one predicted sentences for a single ground truth sentence.  
    \item \textbf{Prune $M$ to remove matches that have non-significant overlap.} To provide credit only when predicted tasks are sufficiently similar to ground truth task sentences, we define a threshold $t$ and remove edges $e \in M$ with $weight(e) < t$. In our work, we set $t = 0.25$. 
\end{enumerate}
\vspace{-2mm}
Fig.~\ref{fig_matching_process} illustrates the matching procedure steps.

\section{Results}
The performance of the various modeling approach configurations are presented in Table \ref{results_table}. The first four rows show the performance of the different SLATE configurations tried and the last row shows the performance of the baseline approach. Our flagship approach is SLATE-NTI which has the following advantages: (1) The highest segmentation performance (B and $\text{B}_{tp}$); (2) Much lower latency compared to the baseline approach; (3) Better or comparable task classification performance with respect to the other approaches. We discuss some observed trends.
\begin{table*}[ht!]
\centering
\small
\scalebox{0.8}{
\begin{tabular}{@{}lllllllllllll@{}}
\toprule
\multicolumn{1}{c}{\multirow{2}{*}{Method}} & \multicolumn{4}{c}{Task (\%)} & \multicolumn{4}{c}{Non-task (\%)} & \multicolumn{1}{c}{\multirow{2}{*}{Acc (\%)}} &
\multicolumn{1}{c}{\multirow{2}{*}{$\text{B}_{tp}$ (\%)}} & \multicolumn{1}{c}{\multirow{2}{*}{B (\%)}} &
\multicolumn{1}{c}{\multirow{2}{1cm}{Latency (ms)}} \\ \cmidrule(lr){2-5}\cmidrule(lr){6-9}
\multicolumn{1}{c}{} & \multicolumn{1}{c}{Rec} & \multicolumn{1}{c}{Prec} & \multicolumn{1}{c}{F1} & \multicolumn{1}{c}{Context Rec} & \multicolumn{1}{c}{Rec} & \multicolumn{1}{c}{Prec} & \multicolumn{1}{c}{F1} & \multicolumn{1}{c}{Context Rec} & \multicolumn{1}{c}{}  & \multicolumn{1}{c}{} & \multicolumn{1}{c}{} & \multicolumn{1}{c}{} \\
\midrule
\begin{tabular}[c]{@{}l@{}}SLATE-NTI with\\ Doc Metadata\end{tabular} & 87.4 & 81.7 & 84.4 & 69.6 & 89.3 & 92.9 & 91.1  & 60.9 & 88.7 & 89.1 & 88.4 &  \multirow{4}{*}{34.2}\\
\cmidrule{1-10}
\begin{tabular}[c]{@{}l@{}}SLATE-BIO with\\ Doc Metadata\end{tabular} & 88.9 & 80.4 & 84.4 & 75.6 & 88.3 & 93.6 & 90.8 & 64.0 & 88.5 & 86.6 & - & \\
\midrule
SLATE-NTI & 90.2 & 81.2 & 85.5 & 78.1 & 88.6 & 94.4 & 91.4 & 59.1 & 89.2 & 84.3 & 83.4 &  \multirow{2}{*}{26.5}\\ 
\cmidrule{1-10}
SLATE-BIO & 87.4 & 83.2 & 85.3 & 70.4 & 90.4 & 93.0 & 91.7 & 63.7 & 89.4 & 83.0 & - &  \\
\midrule
Baseline & 83.1  & 81.5 & 82.3 & 43.0 & 89.8 & 90.1 & 90.2 & 57.8 & 87.4 &  85.5 & 85.3 & 90.6\\
\bottomrule
\end{tabular}
}
\caption{Performance comparison of the various modeling approach configurations on our test set. The classification (recalls, precisions, F1s and accuracy) and segmentation metrics ($\text{B}_{tp}$ and B) are averages over five distinctly seeded training runs for the corresponding method. The latency values were obtained by finding the mean latency of the model inference over each sample in the test set and then performing the average of five such runs.}
\vspace{-3mm}
\label{results_table}
\end{table*}

\vspace{-5mm}
\subsection{The Latency Advantage of SLATE}
As shown in Table \ref{results_table}, the SLATE approaches have 2.6 to 3.4 times lower inference latency compared to the the baseline approach, depending on whether they use document metadata or not. This lower latency of SLATE can be attributed to two main reasons. The first is that SLATE uses a single model for both segmentation and classification, whereas the baseline approach suffers from the combined latency of two separate models. The second reason is that since the classification model of the baseline approach acts on each sentence independently, its inference time scales linearly with the number of sentences in the input. The SLATE approach on the other hand can perform inference on an input containing multiple sentences in a single inference. 

\subsection{SLATE Benefits from Context}
Unlike the sentence classification model in our baseline which has access to only a single sentence per inference, the SLATE approach has access to contextual information since it performs inference on a whole block of text at once. Table \ref{results_table} shows the advantage that access to context gives SLATE compared to the baseline approach: Every SLATE approach has better classification performance (F1 scores and Accuracy) compared to the baseline. To further zoom in on the effect of contextual information, Table \ref{results_table} also shows the recall of the approaches on sentences in the test set that were annotated as being tasks or non-tasks only due to context (see the Context Rec. columns). We see that each of the SLATE approaches has significantly higher recalls compared to the baseline on such sentences. 

\subsection{The Benefit of Multi-task Learning}
SLATE with either the BIO or NTI labeling schemes is inherently a form of multi-task learning as it is trained to both segment text and classify segmented text simultaneously. Still, SLATE-NTI has a stronger multi-task component compared to SLATE-BIO since unlike the BIO scheme, the NTI scheme forces the model to not only learn how to segment out task sentences but non-task sentences as well. Learning how to find the boundaries of non-task sentences is complementary to learning to find boundaries of task sentences. Thus, SLATE-NTI learns to be more effective at segmenting the text compared to SLATE-BIO. This can be seen in Table \ref{results_table} by observing that the $\text{B}_{tp}$ scores for SLATE-NTI configurations are higher than their corresponding SLATE-BIO configurations. 

\subsection{Adapting to Ink using Layout Metadata}
As discussed in Section \ref{domain_adaptation}, we expect that adding document layout information such as line breaks and bullets to the model input should help compensate for the lack of traditional characteristics of natural language such as proper grammar, punctuation, capitalization, verbosity, etc. The results in Table \ref{results_table} substantiate this expectation as we see large margins of improvement (> 3.6\%) in the segmentation metrics (B and $\text{B}_{tp}$) when we compare the SLATE approaches that use document metadata against those that do not. Thus, supplementing the model with document layout information is an effective method to adapt to the segmentation challenges of the inking domain. 

\section{Conclusion}
We have presented SLATE, a single-model, sequence labeling approach for extracting tasks from free-form content. It overcomes ink domain challenges via our custom ink dataset and ink-document layout information. Our flagship configuration, SLATE-NTI, is a single, low-latency model trained for both accurate sentence segmentation and task sentence classification on inked content.



\clearpage

\bibliography{anthology,custom}

\begin{thebibliography}{26}
\expandafter\ifx\csname natexlab\endcsname\relax\def\natexlab#1{#1}\fi

\bibitem[{Apr(2022)}]{April_myscript_api}
 2022.
\newblock Cross-platform handwriting recognition and interactive ink apis.
\newblock \url{https://developer.myscript.com/}.

\bibitem[{Bennett and Carbonell(2005)}]{bennett2005detecting}
Paul~N Bennett and Jaime Carbonell. 2005.
\newblock Detecting action-items in e-mail.
\newblock In \emph{Proceedings of the 28th annual international ACM SIGIR
  conference on Research and development in information retrieval}, pages
  585--586.

\bibitem[{Chen et~al.(2020)Chen, Ruan, Liu, and Lu}]{chen2020seqvat}
Luoxin Chen, Weitong Ruan, Xinyue Liu, and Jianhua Lu. 2020.
\newblock Seqvat: Virtual adversarial training for semi-supervised sequence
  labeling.
\newblock In \emph{Proceedings of the 58th Annual Meeting of the Association
  for Computational Linguistics}, pages 8801--8811.

\bibitem[{Chinchor and Sundheim(1993)}]{muc}
Nancy Chinchor and Beth Sundheim. 1993.
\newblock \href {https://aclanthology.org/M93-1007} {{MUC}-5 evaluation
  metrics}.
\newblock In \emph{Fifth Message Understanding Conference ({MUC}-5):
  Proceedings of a Conference Held in Baltimore, {M}aryland, August 25-27,
  1993}.

\bibitem[{Crawshaw(2020)}]{multitask_learning}
Michael Crawshaw. 2020.
\newblock Multi-task learning with deep neural networks: A survey.
\newblock \emph{arXiv preprint arXiv:2009.09796}.

\bibitem[{Fournier(2013)}]{fournier2013evaluating}
Chris Fournier. 2013.
\newblock Evaluating text segmentation using boundary edit distance.
\newblock In \emph{Proceedings of the 51st Annual Meeting of the Association
  for Computational Linguistics (Volume 1: Long Papers)}, pages 1702--1712.

\bibitem[{Fournier and Inkpen(2012)}]{fournier2012segmentation}
Chris Fournier and Diana Inkpen. 2012.
\newblock Segmentation similarity and agreement.
\newblock \emph{arXiv preprint arXiv:1204.2847}.

\bibitem[{Gericke et~al.(2012)Gericke, Wenzel, Gumienny, Willems, and
  Meinel}]{gericke2012handwriting}
Lutz Gericke, Matthias Wenzel, Raja Gumienny, Christian Willems, and Christoph
  Meinel. 2012.
\newblock Handwriting recognition for a digital whiteboard collaboration
  platform.
\newblock In \emph{2012 International Conference on Collaboration Technologies
  and Systems (CTS)}, pages 226--233. IEEE.

\bibitem[{He et~al.(2020)He, Wang, Wei, Feng, Mao, and
  Jiang}]{seq_label_survey}
Zhiyong He, Zanbo Wang, Wei Wei, Shanshan Feng, Xianling Mao, and Sheng Jiang.
  2020.
\newblock A survey on recent advances in sequence labeling from deep learning
  models.
\newblock \emph{arXiv preprint arXiv:2011.06727}.

\bibitem[{Karp(1980)}]{karp1980algorithm}
Richard~M Karp. 1980.
\newblock An algorithm to solve the m$\times$ n assignment problem in expected
  time o (mn log n).
\newblock \emph{Networks}, 10(2):143--152.

\bibitem[{Keysers et~al.(2016)Keysers, Deselaers, Rowley, Wang, and
  Carbune}]{multilanghandwriting}
Daniel Keysers, Thomas Deselaers, Henry~A. Rowley, Li-Lun Wang, and Victor
  Carbune. 2016.
\newblock \href
  {http://ieeexplore.ieee.org/stamp/stamp.jsp?tp=&arnumber=7478642}
  {Multi-language online handwriting recognition}.
\newblock \emph{IEEE Transactions on Pattern Analysis and Machine
  Intelligence}.

\bibitem[{Le(2020)}]{le2020sequence}
The~Anh Le. 2020.
\newblock Sequence labeling approach to the task of sentence boundary
  detection.
\newblock In \emph{Proceedings of the 4th International Conference on Machine
  Learning and Soft Computing}, pages 144--148.

\bibitem[{Liu et~al.(2019)Liu, Ott, Goyal, Du, Joshi, Chen, Levy, Lewis,
  Zettlemoyer, and Stoyanov}]{liu2019roberta}
Yinhan Liu, Myle Ott, Naman Goyal, Jingfei Du, Mandar Joshi, Danqi Chen, Omer
  Levy, Mike Lewis, Luke Zettlemoyer, and Veselin Stoyanov. 2019.
\newblock Roberta: A robustly optimized bert pretraining approach.
\newblock \emph{arXiv preprint arXiv:1907.11692}.

\bibitem[{Ma et~al.(2018)Ma, Zheng, Xie, Li, Li, and Si}]{token_metrics1}
Chunping Ma, Huafei Zheng, Pengjun Xie, Chen Li, Linlin Li, and Luo Si. 2018.
\newblock Dm\_nlp at semeval-2018 task 8: neural sequence labeling with
  linguistic features.
\newblock In \emph{Proceedings of The 12th International Workshop on Semantic
  Evaluation}, pages 707--711.

\bibitem[{Ramshaw and Marcus(1999)}]{ramshaw1999textchunking}
Lance~A Ramshaw and Mitchell~P Marcus. 1999.
\newblock Text chunking using transformation-based learning.
\newblock In \emph{Natural language processing using very large corpora}, pages
  157--176. Springer.

\bibitem[{Rehbein et~al.(2020)Rehbein, Ruppenhofer, and
  Schmidt}]{rehbein2020improving}
Ines Rehbein, Josef Ruppenhofer, and Thomas Schmidt. 2020.
\newblock Improving sentence boundary detection for spoken language
  transcripts.
\newblock In \emph{Proceedings of the 12th International Conference on Language
  Resources and Evaluation (LREC), May 11-16, 2020, Palais du Pharo, Marseille,
  France}, pages 7102--7111. European Language Resources Association.

\bibitem[{Sang and Buchholz(2000)}]{conllchunking}
Erik~F Sang and Sabine Buchholz. 2000.
\newblock Introduction to the conll-2000 shared task: Chunking.
\newblock \emph{arXiv preprint cs/0009008}.

\bibitem[{Sang and De~Meulder(2003)}]{conll2003}
Erik~F Sang and Fien De~Meulder. 2003.
\newblock Introduction to the conll-2003 shared task: Language-independent
  named entity recognition.
\newblock \emph{arXiv preprint cs/0306050}.

\bibitem[{Segura~Bedmar et~al.(2013)Segura~Bedmar, Mart{\'\i}nez, and
  Herrero~Zazo}]{segura2013semeval}
Isabel Segura~Bedmar, Paloma Mart{\'\i}nez, and Mar{\'\i}a Herrero~Zazo. 2013.
\newblock Semeval-2013 task 9: Extraction of drug-drug interactions from
  biomedical texts (ddiextraction 2013).
\newblock Association for Computational Linguistics.

\bibitem[{Stevenson and Gaizauskas(2000)}]{stevenson2000experiments}
Mark Stevenson and Robert Gaizauskas. 2000.
\newblock Experiments on sentence boundary detection.
\newblock In \emph{Sixth Applied Natural Language Processing Conference}, pages
  84--89.

\bibitem[{Tourille et~al.(2018)Tourille, Doutreligne, Ferret, N{\'e}v{\'e}ol,
  Paris, and Tannier}]{token_metrics2}
Julien Tourille, Matthieu Doutreligne, Olivier Ferret, Aur{\'e}lie
  N{\'e}v{\'e}ol, Nicolas Paris, and Xavier Tannier. 2018.
\newblock Evaluation of a sequence tagging tool for biomedical texts.
\newblock In \emph{proceedings of the Ninth International Workshop on Health
  Text Mining and Information Analysis}, pages 193--203.

\bibitem[{Wang et~al.(2019)Wang, Hosseini, Awadallah, Bennett, and
  Quirk}]{wang2019context}
Wei Wang, Saghar Hosseini, Ahmed~Hassan Awadallah, Paul~N Bennett, and Chris
  Quirk. 2019.
\newblock Context-aware intent identification in email conversations.
\newblock In \emph{Proceedings of the 42nd International ACM SIGIR Conference
  on Research and Development in Information Retrieval}, pages 585--594.

\bibitem[{Wolf et~al.(2020)Wolf, Chaumond, Debut, Sanh, Delangue, Moi, Cistac,
  Funtowicz, Davison, Shleifer et~al.}]{wolf2020transformers}
Thomas Wolf, Julien Chaumond, Lysandre Debut, Victor Sanh, Clement Delangue,
  Anthony Moi, Pierric Cistac, Morgan Funtowicz, Joe Davison, Sam Shleifer,
  et~al. 2020.
\newblock Transformers: State-of-the-art natural language processing.
\newblock In \emph{Proceedings of the 2020 Conference on Empirical Methods in
  Natural Language Processing: System Demonstrations}, pages 38--45.

\bibitem[{Ye et~al.(2005)Ye, Sutanto, Raghupathy, Li, and
  Shilman}]{linegrouping}
Ming Ye, Herry Sutanto, Sashi Raghupathy, Chengyang Li, and Michael Shilman.
  2005.
\newblock Grouping text lines in freeform handwritten notes.
\newblock In \emph{Eighth International Conference on Document Analysis and
  Recognition (ICDAR'05)}, pages 367--371. IEEE.

\bibitem[{Ye and Viola(2004)}]{list_detection}
Ming Ye and Paul Viola. 2004.
\newblock Learning to parse hierarchical lists and outlines using conditional
  random fields.
\newblock In \emph{Ninth International Workshop on Frontiers in Handwriting
  Recognition}, pages 154--159. IEEE.

\bibitem[{Ye et~al.(2007)Ye, Viola, Raghupathy, Sutanto, and
  Li}]{block_grouping}
Ming Ye, Paul Viola, Sashi Raghupathy, Herry Sutanto, and Chengyang Li. 2007.
\newblock Learning to group text lines and regions in freeform handwritten
  notes.
\newblock In \emph{Ninth International Conference on Document Analysis and
  Recognition (ICDAR 2007)}, volume~1, pages 28--32. IEEE.

\end{thebibliography}
\bibliographystyle{acl_natbib}

\appendix

\section{Ink Document Examples}\label{ink_document_examples}

\begin{figure}[t!]
    \centering
    \includegraphics[width=\linewidth]{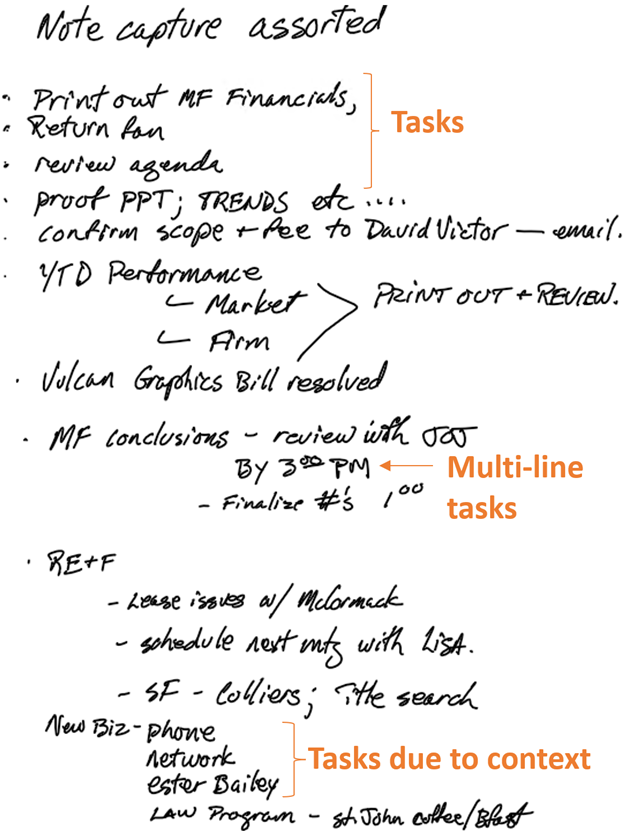}
    \caption{Example of task sentences in an ink document. }
    \label{todo_list_doc}
\end{figure}

Here we provide additional examples of task/non-task sentences occurring in various styles in ink documents. Fig.~\ref{todo_list_doc} is an example of a to-do list style ink document. Here we see task sentences mainly written in the form of  bullets some of which also span over multiple lines. Note certain sentences are tasks only based on the context and might not seem like task sentences otherwise.

\begin{figure}[t!]
    \centering
    \includegraphics[width=\linewidth]{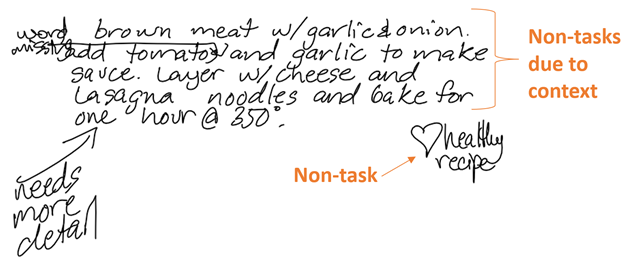}
    \caption{Example of non-task sentences in an ink document. }
    \label{recipe_doc}
\end{figure}

Similarly, Fig.~\ref{recipe_doc} shows an example of an inked recipe content. Although some of the sentences may seem like tasks, they are not in the context of being a recipe. For example, while ``add tomato and garlic to make sauce" may be written like a task sentence, it is not considered a task sentence as it is an instruction in a recipe. 

\section{Token-level to Word-level label Aggregation Rules for Sequence Labeling}
\label{aggregation_rules}

\begin{algorithm}[h!]
    \SetKwInOut{Input}{Input}
    \SetKwInOut{Output}{Output}
    \Input{$tokenLabels$ is a list of a token-level labels for a given word.}
    \Output{Word-level label aggregated from $tokenLabels$.}
    \eIf{`B' in $tokenLabels$}
        {
            return 'B'; 
        }
        {
            return 'I';
        }
    \caption{Sentence-BI rule for aggregating token-level labels to word-level labels.}\label{algorithm1}
\end{algorithm}
\begin{algorithm}[h!]
    \SetKwInOut{Input}{Input}
    \SetKwInOut{Output}{Output}
    \Input{$tokenLabels$ is a list of a token-level labels for a given word.}
    \Output{Word-level label aggregated from $tokenLabels$.}
    \eIf{`B' in $tokenLabels$}
        {
            return 'B'; 
        }
        {
            return $mode(tokenLabels)$;
        }
    \caption{SLATE-BIO rule for aggregating token-level labels to word-level labels.}\label{algorithm2}
\end{algorithm}
\begin{algorithm}[h!]
    \SetKwInOut{Input}{Input}
    \SetKwInOut{Output}{Output}
    \Input{$tokenLabels$ is a list of a token-level labels for a given word.}
    \Output{Word-level label aggregated from $tokenLabels$.}
    \eIf{`N' or `T' in $tokenLabels$}
        {
            \eIf {\# of `T' labels > \# of `N' labels}
            {
                return `T'; 
            }
            {
                return `N'; 
            }
        }
        {
            return `I';
        }
    \caption{SLATE-NTI rule for aggregating token-level labels to word-level labels.}\label{algorithm3}
\end{algorithm}

\section{Training, Implementation and Hyperparameter Details}\label{training_details}
Each of our models were implemented using the HuggingFace Transformers library \cite{wolf2020transformers}. For sequence labeling models, we use the $AutoModelForTokenClassification$ class with the $RoBERTa_{base}$ architecture. For our sentence classification model in the baseline, we use the $AutoModelForSequenceClassification$ class with the $RoBERTa_{base}$ architecture. The model encoders were initialized using the pretrained weights provided by the library. 

For training we use a batch size of 3 and 16 for the sequence labeling models and classification model respectively. All models were fine-tuned on our training set for 100 epochs. The objective for all the models was a class-weighted cross-entropy loss. The learning rate was kept constant at $1\times10^{-6}$ for all of the models. 

While training we use a machine with an NVIDIA RTX 2080ti GPU, Intel i9-9900K CPU and 64GB of RAM. For latency experiments, inference was done on the CPU of the above machine.

\section{Evaluation Procedure}\label{evaluation appendix}

\subsection{Our Evaluation Approach}
Our evaluation procedure has the following steps: 
\begin{enumerate}
    \vspace{-2mm}
    \item Match predicted task sentences to ground truth sentences (task and non-task) that have significant overlap. The matching procedure is described in more detail in Section~\ref{matching_section}. 
    \vspace{-2mm}
    \item Calculate the number of true/false positives/negatives according to the following definitions:
    \vspace{-3mm}
    \paragraph{True Positive:} \hspace{-3mm}Predicted task sentence that is matched to a ground truth task sentence.
    \vspace{-3mm}
    \paragraph{False Positive:} \hspace{-3mm}Predicted task sentence that is matched to a ground truth non-task sentence.
    \vspace{-3mm}
    \paragraph{True Negative:} \hspace{-3mm} Ground truth non-task sentence that is not matched to any predicted task sentence.
    \vspace{-3mm}
    \paragraph{False Negative:} \hspace{-3mm}Ground truth task sentence not matched to any predicted task sentence.\vspace{2mm} \\
    Fig.~\ref{fig_intermediate_metrics} shows the calculation of true/false positives/negatives for a sample input text, prediction and ground truth annotation. 
    \vspace{-2mm}
    \item Calculate standard classification metrics (accuracy, recall, precision, f1-scores, etc.) from the true/false positives/negatives to understand the task classification performance. When calculating non-task recall/precision, we treat true/false positives as true/false negatives and vice versa. 
    \vspace{-2mm}
    \item Calculate sentence segmentation metrics to evaluate the quality of the extracted task sentences. In this work we use the \textit{boundary similarity (B)} sentence segmentation metric introduced in \cite{fournier2013evaluating}. The boundary similarity metric is based on the \textit{boundary edit distance (BED)} introduced in \cite{fournier2012segmentation}. Concisely, the boundary similarity metric penalizes a predicted segmentation based on the number of edits required to transform the predicted segmentation to the ground truth segmentation. Near boundary misses are penalized less compared to full boundary misses/additions. \textit{B} is a score from 0-1 where a higher score represents a better predicted segmentation and a 1 represents a perfect match to the ground truth segmentation. More metric details are in Appendix~\ref{app_boundary_sim}.
    
    In our application, since the segmentation quality of extracted task sentences matters more than that of non-task sentences, we also compute a modified version of this metric which we call the \textit{true positive boundary similarity ($B_{tp}$)}. The formula to compute $B_{tp}$ is the same as Equation \ref{eq_boundary_similarity} (Appendix~\ref{app_boundary_sim}) except that in the segmentations that we compare, we only include the boundaries of true positive tasks in the predicted segmentation and the corresponding boundaries for the matched tasks in the ground truth segmentation. 
\end{enumerate}

\subsection{Why not use standard NER Evaluation instead?}

An alternative approach could be to leverage Named Entity Recognition (NER) metrics where the entities we are trying to recognize are task sentences. But these metrics are not without issues either. NER systems are typically evaluated by calculating precision, recall, and F1-scores at either the token level \cite{token_metrics1, token_metrics2} or at the entity level \cite{conll2003, segura2013semeval}. Token-level metrics suffer from being difficult to interpret compared to entity-level metrics. However, entity-level metrics are often too strict, giving credit only when predicted entities match the ground truth exactly \cite{conll2003}. For example, in our scenario, if our model misses only a single token in an extracted task sentence, it would get no credit.

To address this, other evaluation schemes that provide credit for partial entity matches have been proposed \cite{segura2013semeval, muc}, but these tend to be more complex and difficult to interpret compared to the standard sentence classification metrics. In this work, we propose an evaluation procedure that allows us to calculate metrics that can be interpreted in the same way that standard sentence classification and segmentation metrics are, but at the same time provides enough slack to allow partial matches of predicted and ground truth task sentences.

\subsection{The Boundary Similarity (B) Metric}\label{app_boundary_sim}

Let us define a segmentation $S$ of text to be a sequence of boundary positions where each boundary represents where a sentence begins and/or ends. A boundary can be placed between words and by default we place boundaries before the first word and after the last word in the text. Boundary edit distance $(BED)$ is a measure of the minimum number of edits that need to be made to a given segmentation $S_1$ to make it identical to another segmentation $S_2$. There are three types of edit operations we can make to $S_1$ in order to bring it into parity with $S_2$:
    \begin{itemize}
        \item \textbf{Addition (A):} When $S_1$ is missing a boundary that is in $S_2$ we can add a boundary to $S_1$. 
        \item \textbf{Deletion (D):} When $S_1$ has a boundary where $S_2$ does not, we can delete this boundary from $S_1$.
        \item \textbf{$n$-wise Transposition (T):} When $S_1$ misses a boundary in $S_2$ but has one in the near neighborhood, instead of making two edits to $S_1$ (one A and one D operation), we allow a single T operation which involves transposing/shifting the near boundary in $S_1$ to the corresponding position in $S2$. The parameter $n$ determines how far boundaries in $S_1$ and $S_2$ can be to be considered for a T operation instead of an A or D operation. In this work, we set $n = 2$, allowing transpositions when boundaries differ by a maximum of 2 positions.
    \end{itemize}
        
    Suppose we calculate the $BED$ for two segmentations of the same text $S_1$ and $S_2$. The $BED$ outputs the following: $N_M$ is the number of perfect matches between boundaries, requiring no edits; $N_A$ is the number of A operations required; $N_D$ is the number of D operations required; the set $S_t = \{t~|~t~=~\#~\text{positions a boundary should be shifted} \\ \forall~T~\text{operations required}\}$. Then \textit{boundary similarity (B)} between $S_1$ and $S_2$ is calculated as follows: 
    \begin{equation}
        B(S_1, S_2) = 1 - \frac{N_A + N_D + \sum_{t \in S_t}\frac{t}{n}}{N_M + N_A + N_D + |S_t|}
        \label{eq_boundary_similarity}
    \end{equation}
    Essentially, $B$ gives no credit when a boundary is completely missed (A or D operation required) but gives partial credit when a near miss occurs (T operation required). For a more detailed explanation of this metric you may refer to \cite{fournier2013evaluating}.
    
\section{Limitations}\label{limitations}
Here we discuss limitations of the work. First, since there are no past works and open datasets in the literature for our task, we are unable to benchmark against past works directly. To help address this, we have decided to open-source our dataset, modeling code, and evaluation code, so future research works can leverage these for benchmarking purposes. Still, this dataset is not very large. It consists of roughly 200 annotated ink documents. While this gave us a decent number of task and non-task sentences for fine-tuning a pretrained model and evaluating it on our task, with more data there are other approaches that we could take to further supplement our approach. For example, while we do try to address domain-specific noise such as errors introduced by handwriting recognition or poor grammar in inked content using document layout information, with a larger ink document corpus, we could try supplementing our methodology for domain adaptation with language modeling. Finally, while we work with free-form content like inked documents, our work assumes the input to be recognized text from this content rather than the raw content (ink strokes). For example, handwriting recognition and document layout analysis methods are out of scope for this work. We cite examples of other works in literature and APIs that deal with these components in Sections 2 and 3.1.

\end{document}